\pdfoutput=1

\documentclass[11pt,table]{article}

\usepackage[final]{acl}

\usepackage{times}
\usepackage{latexsym}

\usepackage[T1]{fontenc}

\usepackage[utf8]{inputenc}

\usepackage{microtype}

\usepackage{inconsolata}

\usepackage{graphicx}

\usepackage{booktabs}  
\usepackage{epigraph}  
\usepackage{algorithm} 
\usepackage{algpseudocode} 
\usepackage{amsmath} 

\usepackage{xcolor}
\usepackage[most]{tcolorbox}
\usepackage{fontawesome5}
\usepackage{enumitem}
\usepackage{float}

\definecolor{boxTitleBlue}{RGB}{0, 80, 100}
\definecolor{boxWrongRed}{RGB}{204, 0, 0}
\definecolor{boxCorrectGreen}{RGB}{0, 128, 0}
\definecolor{feedbackYellow}{RGB}{255, 250, 225}
\newcommand{\boxheading}[1]{\par\noindent\vspace{3mm}\textcolor{boxTitleBlue}{\textbf{#1:}}}

\definecolor{promptpurple}{RGB}{128, 80, 180}

\newtcolorbox{promptbox}[2][]{
    enhanced,
    colback=white,             
    coltext=black,             
    colframe=promptpurple,     
    colbacktitle=promptpurple, 
    coltitle=white,            
    fonttitle=\bfseries\sffamily, 
    fontupper=\sffamily,       
    boxrule=0pt,               
    arc=3mm,                   
    title=#2,                  
    #1
}

\title{Retrieval Enhanced Feedback via In-context Neural Error-book}

\author{Jongyeop Hyun \\
 School of CSE\\
 Chung-Ang University\\
  \texttt{jesussuej@cau.ac.kr}\thanks{contact: mldljyh@postech.ac.kr} \\
  \\\And
 Bumsoo Kim\thanks{Corresponding author.} \\
 School of CSE\\
 Chung-Ang University\\
 \texttt{bumsoo@cau.ac.kr} \\}

\begin{document}
\maketitle
\begin{abstract}
Recent advancements in Large Language Models (LLMs) have significantly improved reasoning capabilities, with in-context learning (ICL) emerging as a key technique for adaptation without retraining.
While previous works have focused on leveraging correct examples, recent research highlights the importance of learning from errors to enhance performance.
However, existing methods lack a structured framework for analyzing and mitigating errors, particularly in Multimodal Large Language Models (MLLMs), where integrating visual and textual inputs adds complexity.
To address this issue, we propose REFINE: Retrieval-Enhanced Feedback via In-context Neural Error-book, a teacher-student framework that systematically structures errors and provides targeted feedback.
REFINE introduces three systematic queries to construct structured feedback---Feed-Target, Feed-Check, and Feed-Path---to enhance multimodal reasoning by prioritizing relevant visual information, diagnosing critical failure points, and formulating corrective actions.
Unlike prior approaches that rely on redundant retrievals, REFINE optimizes structured feedback retrieval, improving inference efficiency, token usage, and scalability.
Our results demonstrate substantial speedup, reduced computational costs, and successful generalization, highlighting REFINE’s potential for enhancing multimodal reasoning.
\end{abstract}
\section{Introduction}
\label{sec:intro}

\setlength\epigraphwidth{\columnwidth}
\setlength\epigraphrule{0pt}
\epigraph{\textit{``The only real mistake is the one from which we learn nothing.''} --- Henry Ford}


Recent LLM advancements show superior reasoning performance, with extensive research on in-context learning (ICL) to enhance human-like reasoning capabilities.
Early works on ICL generated responses based on a few provided correct examples, enabling the models to adapt to new tasks without extensive retraining~\citep{brown.etal_2020, dong-etal-2024-survey, wei.etal_2023a}.
Although ICL has demonstrated effectiveness by primarily leveraging \textit{correct} examples~\citep{min-etal-2022-rethinking}, subsequent works have re-examined the fact that human learning is also deeply rooted in learning from \textit{errors}~\citep{Edmondson1996LearningFM,chialvo.bak_1999, Edmondson1999PsychologicalSA}.
Recent studies suggest that incorporating errors into the learning process can further improve LLM performance~\citep{sun-etal-2024-retrieved}.
These approaches typically identify recurring errors, extract underlying principles from them, and apply these insights to prevent similar errors in the future.

However, a critical limitation of these methodologies is the absence of a systematic framework for structuring errors, making it difficult to analyze and mitigate failure cases effectively.
This challenge becomes even more pronounced in Multimodal Large Language Models (MLLMs) that jointly process multimodal inputs~\citep{zhao.etal_2023a}.
Unlike unimodal LLMs where errors in textual reasoning can often be traced and corrected through established interpretability techniques, MLLMs introduce additional complexity due to the integration of visual and textual modalities.
Without a structured approach to diagnosing and addressing previous errors, failures in one modality can propagate through the system, making it more difficult to ensure reliable and interpretable outcomes~\citep{lau2025uncertainty}.
While recent works such as multimodal-CoT have shown that incorporating textual CoT reasoning and in-context learning can improve performance, a fundamental gap persists: the lack of structured error analysis for multimodal reasoning makes it unclear whether existing MLLMs can fully leverage CoT reasoning for visual understanding~\citep{alayrac.etal_2022a, zhao.etal_2023a}.
Furthermore, the interpretability of MLLMs remains a significant challenge, as current approaches do not adequately explain how visual information contributes to reasoning, highlighting the need for a more rigorous framework for structuring and mitigating errors in multimodal AI systems.

To address these challenges, we propose \textbf{REFINE: Retrieval-Enhanced Feedback via In-context Neural Error-book}.
Our proposed REFINE is a teacher-student framework where the teacher generates a structured Error-book based on the student's observed errors, establishes question-level feedback based on this Error-book, and the student retrieves and applies proper feedback to prevent the recurrence of similar errors.
Instead of expecting the MLLM to extract the optimal intuition to resolve the error from the initial response, we prompt it to structure what went wrong during the inference process.
In MLLMs, effective reasoning requires a stronger focus on visual inputs.
To enhance this process, we introduce three structured feedback mechanisms: Feed-Target, Feed-Check, and Feed-Path.

``Feed-Target'' extracts high-level observations essential for accurate inference based on an image-question pair.
For instance, when answering a question that requires counting pedestrians or vehicles, the model must prioritize proper object detection within the visual input.
``Feed-Check'' retrospectively analyzes errors, identifying the most critical failure points.
For example, in an image-question-answer triplet, an error may stem from incorrect perception of the object \textit{people}.
Thirdly, ``Feed-Path'' formulates corrective actions by generating explicit instructions to refine the model’s response and mitigate previously identified errors.
We integrate these feedback mechanisms while excluding self-regulatory feedback, which our analysis shows introduces noise rather than improving response quality. Since our feedback structure is well-organized, storing multiple insights for similar questions is unnecessary. Unlike previous approaches that often rely on retrieving and processing multiple samples, REFINE employs a deterministic single-nearest-neighbor strategy for its structured feedback. This ensures consistency and low overhead at inference time, in stark contrast to the inefficiencies and stochastic behavior of traditional ICL approaches, thereby enabling a structured framework to infer the correct chain-of-thought reasoning without redundant retrievals in multimodal tasks. Additionally, we demonstrate that task-level insight retrieval offers no measurable benefit in multimodal question-answering benchmarks.

Besides accuracy, our method significantly outperformed baselines in terms of inference efficiency.
Our structured feedback retrieval is substantially faster ($44.7-76.4\times$ speedup compared to the RICP baseline) and more token-efficient, improving spatial complexity (approximately 64.2\% fewer tokens).
The use of precomputed embeddings and the removal of clustering significantly reduce computational costs, demonstrating scalability and feasibility in real-time settings.
Furthermore, successful generalization from smaller subsets (MME-RealWorld-Lite) to larger-scale tests (MME-RealWorld) clearly illustrates the practical scalability of our approach.

\section{Related Work}
\label{sec:related_work}

\subsection{Chain-of-Thought Reasoning}
Chain-of-Thought (CoT) reasoning enhances Large Language Model (LLM) problem-solving by breaking down tasks into intermediate logical steps, akin to human cognition~\citep{wei2022chain}. This structured approach has improved performance in mathematical, commonsense, and multimodal tasks. However, CoT's efficacy depends on the model's inherent reasoning and prompt quality. A key challenge is error propagation from incorrect intermediate steps, necessitating refinements like retrieval or error-correction mechanisms for improved robustness~\citep{cao.etal_2023}.

\subsection{Language Instruction Understanding in Multimodal Tasks}
Language instructions are vital for AI-user interaction in multimodal tasks, traditionally requiring precision. While recent Multimodal Large Language Models (MLLMs) better handle complex instructions, ambiguity remains a challenge~\citep{fu.etal_2023, fei-etal-2022-cqg}. Efforts like WAFFLECLIP~\citep{roth2023waffling} and FUDD~\citep{esfandiarpoor2024follow} address polysemy in image classification, and REPHRASE~\citep{prasad2024rephrase} uses iterative prompting for visual question answering. However, many existing methods demand extensive interactions or predefined rules, hindering generalization and scalability across diverse tasks and models.

\subsection{Complex Reasoning with Multimodal Large Language Models}
MLLMs increasingly integrate visual and textual information~\citep{caffagni-etal-2024-revolution}, yet comprehending complex language in visual contexts remains difficult~\citep{zhao.etal_2023a, NEURIPS2024_32923dff}. Approaches to enhance MLLM reasoning are either training-based, aligning models with image-text data, sometimes using synthetic data~\citep{davidson.etal_2025}, or non-training-based, like CoT, which simulates step-by-step reasoning~\citep{NEURIPS2024_d74033a2, NEURIPS2024_32923dff}. Training-based methods can be data-intensive and costly~\citep{davidson.etal_2025}, while non-training methods often presume strong pre-existing reasoning capabilities~\citep{zhang.etal_2024k}, limiting general-purpose intelligence development.

\subsection{Systematic Structuring of MLLMs}
Systematically structuring MLLMs with Retrieval-Augmented Generation (RAG)~\citep{lewis2020retrieval} is crucial for performance and scalability. Studies like~\citet{wang-etal-2024-searching} and VisRAG~\citep{yu.etal_2024} show RAG's benefits in multimodal contexts, while RAVEN~\citep{rao.etal_2024} highlights gains in tasks like image captioning. This underscores the need for integrating retrieval modules for adaptability. Our work also advocates for refined feedback flows, drawing on structured techniques to filter unhelpful signals, akin to our concise ``Error-book'' concept. Existing methods, however, often face data curation challenges or domain-specific limitations. Our approach aims for a distilled feedback structure, enhancing MLLM resilience and extensibility with minimal overhead.

\subsection{Advancements in Error-Driven Learning}
Recent error-driven learning methods include LEAP~\citep{zhang.etal_2024i}, generating static principles from errors, and TRAN~\citep{tong.etal_2024a}, maintaining rules to avoid past mistakes. RICP~\citep{sun-etal-2024-retrieved} clusters errors into task-level principles and retrieves question-level insights. These methods often rely on generalized principles or pre-clustered errors, potentially misaligning with immediate task goals. Our approach differs by restructuring feedback based on the \textbf{Feedback Model}~\citep{hattie.timperley_2007}, focusing on task/process-level guidance through an ``Error-book'' addressing: \textbf{Feed-Target} (goal), \textbf{Feed-Check} (progress), and \textbf{Feed-Path} (actions). This prioritizes task-specific guidance, enabling efficient, deterministic retrieval without teacher intervention during inference, thus reducing overhead and improving precision over methods with rigid clustering or insufficient task focus.
\section{Method}
\label{sec:method}

\begin{figure*}[t!]
     \centering
     \includegraphics[width=\textwidth]{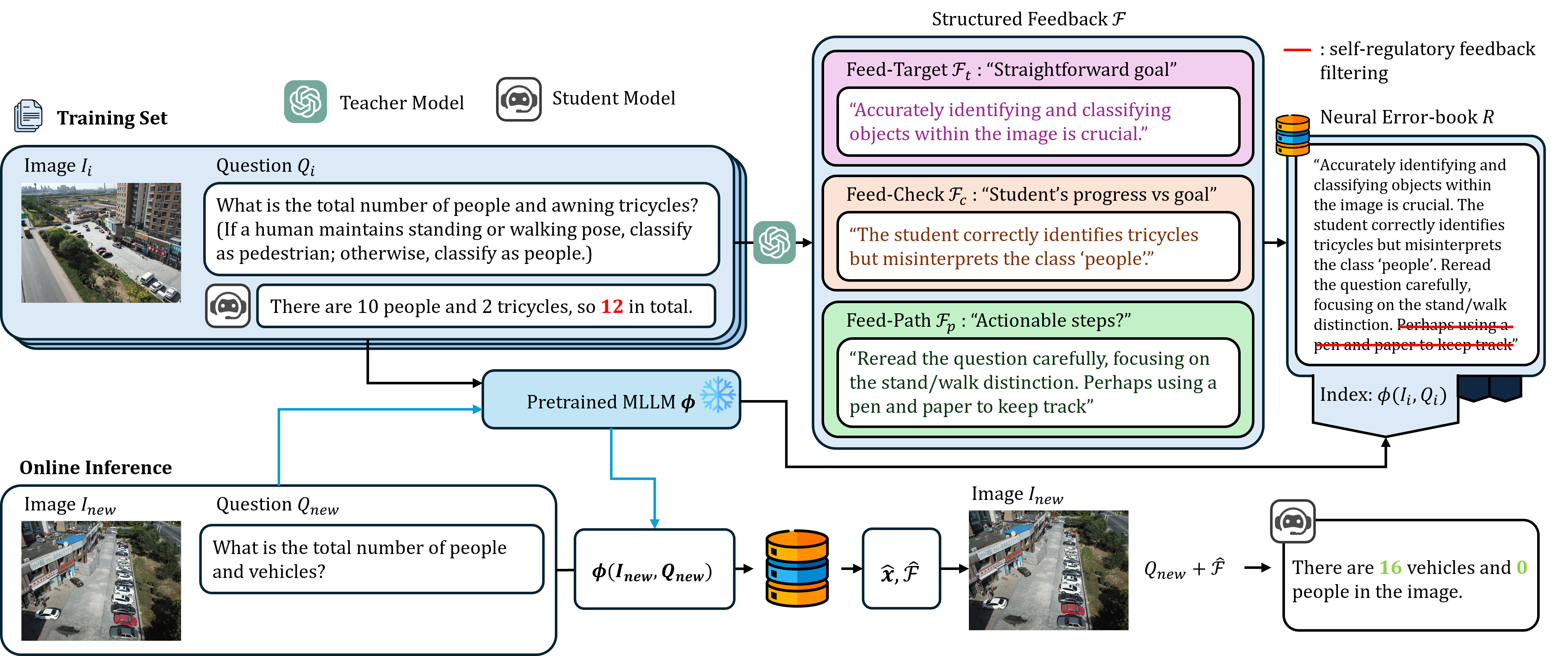}
     \caption{Overview of the REFINE. Given training set image-question pairs, REFINE extracts structured feedbacks under three systematic queries: Feed-Target, Feed-Check and Feed-Path. The self-regulatory questions are filtered out to construct the final feedback and our final Neural Error-book $R$ is indexed under the multimodal embedding via the pre-trained model $\phi$. During inference, the input image and question pairs are embedded to retrieve the most similar feedback within $R$ under their multimodal embedding as the index. The retrieved feedback $\hat{\mathcal{F}}$ is used to enhance the query and obtain the final result.}
     \label{fig:figure2}
\end{figure*}

The objective of this paper is to develop a structured Neural Error-book that systematically provides feedback to enhance model performance effectively.
To achieve this, the Neural Error-book construction involves three distinct feedback formulation stages.
The detailed pseudocode for our REFINE framework is presented in Appendix~\ref{sec:appendix_algorithm}.

\subsection{Structured Feedback Generation}
REFINE is a teacher-student framework where the teacher model systematically analyzes errors made by the student model.
We designed three guiding principles to construct a final structured feedback for each error (student misprediction) inspired by classic educational psychology~\citep{hattie.timperley_2007}: Feed-\textit{Target}, Feed-\textit{Check}, and Feed-\textit{Path}.
These principles guide structured error analysis and facilitate the generation of precise, actionable insights.
Given a multimodal QA benchmark, our process begins by evaluating the student model on a training set of image-question pairs $\{(I_i, Q_i)\}_{i=1}^N$. Errors from student predictions $\hat{A}_i$ (incorrect answers) are aggregated and provided to the teacher model alongside their ground-truth reference $A_i$. The teacher model generates structured feedback through the threefold analysis:

\begin{enumerate}
    \item \textbf{Feed-Target}: \textit{``What is the straightforward goal of this task?''} Clarifies essential task requirements by extracting high-level observations necessary for accurate inference (e.g., ``Proper object detection is essential for counting pedestrians and vehicles'').
    
    \item \textbf{Feed-Check}: \textit{``How does the student's current progress align with the goal?''} Analyzes the student's mispredictions retrospectively, pinpointing critical failures in perception or reasoning (e.g., ``Misclassification of `people' due to overlooking pose criteria'').

    \item \textbf{Feed-Path}: \textit{``What actionable steps bridge the gap to achieve the goal?''} Formulates explicit corrective instructions designed to help prevent error recurrence (e.g., ``Re-analyze image regions with sitting figures using the question's pose definitions'').
\end{enumerate}

\subsection{Feedback Filtering}
\label{sec:feedback_filtering}
After Feed-Target, Feed-Check and Feed-Path, the resulting feedback instances are categorized. This classification is performed automatically by our teacher model. The model is provided with the definitions of ``Task/Process-relative'' feedback (that directly corrects task-specific errors or adds specifics for reasoning, e.g., ``Adjust counts for occluded objects.'') and ``Self-Regulatory'' feedback (that addresses metacognitive habits or personal traits, e.g., ``To improve accuracy, try solving similar problems multiple times.'') from ~\citet{hattie.timperley_2007} and then prompted to classify each generated instance. This automated method ensures the process is consistent and reproducible. Based on empirical observations suggesting that self-regulatory feedback tends to hinder final Chain-of-Thought (CoT) performance, we filter out feedback classified as self-regulatory.



\subsection{Neural Error-book Construction}
After filtering the generated feedback to retain only actionable task/process-level feedback (denoted as $\{\mathcal{F}_i\}_{i\in N_e}$, where $N_e$ denotes error cases), we pair each feedback with the joint embedding for the corresponding image-question pairs to construct a structured Neural Error-book database $R$.
Given $\phi$ as a pre-trained embedding model (e.g., voyage-multimodal-3~\citep{voyage_2024d}) that obtains a joint multimodal embedding from each image-question pair in the training set $x_i = (I_i, Q_i)$, we construct a Neural Error-book $R$ as:
\begin{equation}
    R=\{\phi(x_i),\mathcal{F}_i\}_{i\in N_e}
\end{equation}
Since the Error-book is indexed with the joint embedding of the image-question pair, it enables efficient retrieval at inference time.
Note that unlike previous work, we store only a single well-structured feedback instead of storing multiple redundant insights and clustering them afterwards for similar questions.

\subsection{REFINE}
During inference on unseen image-question pairs, we leverage the constructed Neural Error-book, which stores structured feedback indexed by the multimodal joint embeddings of the training set’s image-question pairs.
The retrieved feedback enhances the in-context feedback for the student model, improving precision and overall result quality.
Given an unseen query $x_{\text{query}} = (I_{\text{query}}, Q_{\text{query}})$, our REFINE framework first computes the multimodal embedding of the query $\phi(x_{\text{query}})$.
Then, from the Neural Error-book $R$, we retrieve the most relevant image-text sample in the training set to obtain the corresponding structured feedback $\hat{\mathcal{F}}$.
\begin{equation}
    \hat{x}, \hat{\mathcal{F}} = \arg\max_{(\phi(x_i), \mathcal{F}_i)\in R} \frac{\phi(x_{\text{query}}) \cdot \phi(x_i)}{\|\phi(x_{\text{query}})\|\|\phi(x_i)\|}
\end{equation}

The retrieved feedback $\hat{\mathcal{F}}$ is systematically integrated into the student model's prompt to guide its reasoning. Specifically, $\hat{\mathcal{F}}$ is appended to the original question $Q_{\text{query}}$, forming an enhanced prompt $P_{\text{enhanced}}$ that combines task context with actionable corrective instructions:  
\begin{equation}
    P_{\text{enhanced}} = \langle Q_{\text{query}}, \hat{\mathcal{F}} \rangle
\end{equation}
where $\langle \cdot \rangle$ denotes the structured prompt format. This design ensures the student model processes both the question and feedback in a unified context, directing attention to previously overlooked criteria or reasoning steps.

This deterministic single-nearest-neighbor strategy ensures consistency and low overhead at inference time, in stark contrast to the inefficiencies and stochastic behavior of traditional ICL approaches.


\section{Experiment}
\label{sec:experiment}


%
\subsection{Experimental Setup}

We evaluated the effectiveness of our feedback mechanisms using three multimodal reasoning benchmarks: \textbf{MME-RealWorld}~\citep{zhang.etal_2024j}, \textbf{MMStar}~\citep{chen.etal_2024c}, and \textbf{SEED-Bench-2-Plus}~\citep{li.etal_2024d}. These benchmarks collectively emphasize diverse multimodal reasoning capabilities, including error diagnosis, procedural correction, and text-rich visual comprehension. The evaluation framework employed for these benchmarks was VLMEvalKit~\citep{duan2024vlmevalkit}. All models in our experiments were used with a temperature setting of 0.0, and results were reported using the pass@1 metric.
Specifically:

\begin{itemize}
\item \textbf{MME-RealWorld} covers complex visual reasoning tasks derived from realistic applications across diverse domains (e.g., Autonomous Driving, Diagrams, OCR).
\item \textbf{MMStar} explicitly selects reasoning problems requiring multimodal integration, filtering out problems solvable without visual context.
\item \textbf{SEED-Bench-2-Plus} evaluates text-rich visual comprehension through 2.3K multiple-choice questions spanning Charts, Maps, and Webs. Its scenarios simulate real-world complexity with embedded textual elements, assessing models' capacity to interpret visually grounded textual information.
\end{itemize}
To verify generalizability and scalability across different model sizes, we selected two representative multimodal models:
\begin{itemize}
    \item \textbf{Pixtral-12B}~\citep{agrawal2024pixtral12b} (a high-capacity model)
    \item \textbf{Qwen2.5-VL-3B-Instruct}~\citep{bai2025qwen25vltechnicalreport} (a compact model)
\end{itemize}

Feedback generation employed Gemini-1.5-Pro~\citep{geminiteam2024gemini15unlockingmultimodal} for MME-RealWorld and MMStar, while SEED-Bench-2-Plus utilized Gemini-2.0-Flash~\citep{gemini2.0flash_2024e} due to its specialized text-rich processing capabilities. \\ 
We constructed our training and evaluation datasets as follows:
\begin{itemize}
    \item \textbf{MME-RealWorld}: After creating an Error-book based on MME-RealWorld-Lite (Reasoning) subset, we applied it directly to the remaining MME-RealWorld Reasoning subset, excluding the Lite portion to rigorously test generalization.
    \item \textbf{MMStar}: We divided the total items in the Instance Reasoning and Logical Reasoning categories into two equal parts, with one part forming the Train Set for creating the Error-book and the remaining items forming a separate Test Set.
    \item \textbf{SEED-Bench-2-Plus}: We utilized the full benchmark dataset (2.3K items) and split it into equal halves, establishing distinct Train and Test sets to evaluate model adaptation to text-rich scenarios while preventing data leakage.
\end{itemize}
Our structured feedback approach was compared against the following baseline feedback methods:
\begin{itemize}
    \item \textbf{Standard Prompting}~\citep{brown.etal_2020}: The LLM is asked to output the answer directly, without the intermediate reasoning process.
    \item \textbf{Chain of Thought}~\citep{wei2022chain}: The LLM is instructed to think step-by-step before providing the answer.
    \item \textbf{Direct Feedback}~\citep{daheim-etal-2024-stepwise}: unstructured, open-ended feedback without explicit instructional framing.
    \item \textbf{RICP}~\citep{sun-etal-2024-retrieved}: retrieved principles from errors clustered by error type, providing task- and question-specific guidance.
\end{itemize}

\subsection{Results and Analysis}


\begin{table*}[ht]
\centering
\resizebox{\textwidth}{!}{
\begin{tabular}{@{}lccccc@{}}
\toprule
\textbf{Method} & \multicolumn{5}{c}{\textbf{MME-RealWorld (Reasoning)}} \\
\cmidrule(lr){2-6}
& \textbf{\shortstack{Autonomous\\Driving}} & \textbf{\shortstack{Diagram\\ \& Table}} & \textbf{Monitoring} & \textbf{\shortstack{OCR with\\Complex Context}} & \textbf{Overall} \\
\midrule
\rowcolor[gray]{0.9}\multicolumn{6}{c}{\textbf{Pixtral-12B}} \\ 
Standard Prompting & 29.66 & 27.75 & 15.80 & 34.00 & 27.82 \\
CoT & 27.86 & 28.00 & 22.70 & 44.25 & 30.16 \\
Direct Feedback & 33.05 & 32.00 & 24.71 & 40.50 & 32.89 \\
RICP & 29.87 & 28.25 & 23.56 & 44.25 & 31.26 \\
\textbf{REFINE} & \textbf{35.06} {\footnotesize(+5.40)} & \textbf{51.00} {\footnotesize(+23.25)} & \textbf{31.32} {\footnotesize(+15.52)} & \textbf{58.25} {\footnotesize(+24.25)} & \textbf{41.92} {\footnotesize(+14.10)} \\
\midrule
\rowcolor[gray]{0.9}\multicolumn{6}{c}{\textbf{Qwen2.5-VL-3B-Instruct}} \\ 
Standard Prompting & 21.19 & 21.75 & 14.94 & 40.25 & 23.90 \\
CoT & 27.54 & 25.50 & 19.25 & \textbf{41.75} & 28.49 \\
Direct Feedback & 29.98 & 19.75 & 21.55 & 35.00 & 27.58 \\
RICP & 19.60 & 23.00 & 13.79 & 39.75 & 23.14 \\
\textbf{REFINE} & \textbf{34.43} {\footnotesize(+13.24)} & \textbf{26.75} {\footnotesize(+5.00)} & \textbf{24.71} {\footnotesize(+9.77)} & 38.50 {\footnotesize(-1.75)} & \textbf{32.12} {\footnotesize(+8.22)} \\
\bottomrule
\end{tabular}
}
\caption{Performance (Accuracy \%) of REFINE against baseline methods on MME-RealWorld (Reasoning) sub-tasks. Parentheses for REFINE indicate improvement over Standard Prompting.}
\label{tab:results}
\end{table*}

\begin{table*}[ht]
    \centering
    \resizebox{\textwidth}{!}{
    \begin{tabular}{@{}lccccccc@{}}
    \toprule
    \textbf{Method} & \multicolumn{3}{c}{\textbf{MMStar (Reasoning)}} & \multicolumn{4}{c}{\textbf{SEEDBench-2-Plus}} \\
    \cmidrule(lr){2-4} \cmidrule(lr){5-8}
    & \textbf{Instance} & \textbf{Logical} & \textbf{Overall} & \textbf{Chart} & \textbf{Map} & \textbf{Web} & \textbf{Total} \\
    \midrule
    \rowcolor[gray]{0.9}\multicolumn{8}{c}{\textbf{Pixtral-12B}} \\
    Standard Prompting & 59.2 & 48.8 & 54.0 & 50.86 & 50.00 & 56.67 & 52.23 \\
    CoT & 59.2 & \textbf{53.6} & 56.4 & 49.14 & 47.04 & 54.55 & 49.96 \\
    Direct Feedback & 60.8 & 42.4 & 51.6 & 47.65 & 47.29 & 50.61 & 48.38 \\
    RICP & 59.2 & 48.0 & 53.6 & 49.63 & 46.55 & 47.27 & 47.85 \\
    \textbf{REFINE} & \textbf{64.8} {\footnotesize(+5.6)} & 50.4 {\footnotesize(+1.6)} & \textbf{57.6} {\footnotesize(+3.6)} & \textbf{56.30} {\footnotesize(+5.44)} & \textbf{52.46} {\footnotesize(+2.46)} & \textbf{57.88} {\footnotesize(+1.21)} & \textbf{55.39} {\footnotesize(+3.16)} \\
    \midrule
    \rowcolor[gray]{0.9}\multicolumn{8}{c}{\textbf{Qwen2.5-VL-3B-Instruct}} \\
    Standard Prompting & 64.0 & 52.0 & 58.0 & \textbf{63.57} & 53.48 & 78.48 & 64.33 \\
    CoT & 56.0 & \textbf{54.4} & 55.2 & 57.46 & 47.51 & 76.97 & 59.60 \\
    Direct Feedback & 63.2 & 52.0 & 57.6 & 61.86 & 52.99 & \textbf{78.79} & 63.63 \\
    RICP & 62.4 & 52.8 & 57.6 & 62.59 & 53.23 & 73.33 & 62.40 \\
    \textbf{REFINE} & \textbf{69.6} {\footnotesize(+5.6)} & 52.0 {\footnotesize(+0.0)} & \textbf{60.8} {\footnotesize(+2.8)} & 62.84 {\footnotesize(-0.73)} & \textbf{54.98} {\footnotesize(+1.50)} & 78.18 {\footnotesize(-0.30)} & \textbf{64.50} {\footnotesize(+0.17)} \\
    \bottomrule
    \end{tabular}
    } 
    \caption{Performance (Accuracy \%) of REFINE against baseline methods on MMStar (Reasoning) and SEEDBench-2-Plus benchmarks. Parentheses for REFINE indicate improvement over Standard Prompting.}
   \label{tab:combined_results}
\end{table*}

\paragraph{Structured Feedback Outperforms Clustering and Unstructured Approaches}
The superiority of our method over RICP and Direct Feedback highlights the importance of \textbf{task-specific granularity} over cluster-level generalizations. For instance, in MME-RealWorld(Reasoning)'s OCR with Complex Context, Pixtral-12B achieves a 24.25-point gain over Standard Prompting with our method, compared to RICP's marginal improvements. This suggests that cluster-level principles fail to address nuanced errors like misclassifying dynamic objects (e.g., distinguishing ``people'' vs. ``pedestrians'' based on pose). Our structured feedback, which explicitly defines task goals (Feed-Target) and actionable corrections (Feed-Path), bridges this gap by contextualizing errors within the specific reasoning process required for the task.

\paragraph{Domain-specific Feedback Efficacy}
Performance gains vary considerably across domains, revealing feedback-task alignment dynamics:
\begin{itemize}
    \item \textbf{Diagram and Table Interpretation (MME-RealWorld)}: The significant improvement (+23.25 for Pixtral-12B) stems from feedback that clarifies hierarchical relationships (e.g., ``Focus on nested chart labels first'')---critical for parsing complex diagrams.
    \item \textbf{Logical Reasoning (MMStar)}: Smaller gains (+1.6 for Pixtral-12B) suggest feedback is less effective for abstract reasoning requiring implicit world knowledge (e.g., causality). Here, structured feedback aids factual corrections (e.g., misidentified object relationships) but struggles with higher-order logic gaps.
    \item \textbf{OCR Decline in Smaller Models}: Qwen2.5-VL-3B-Instruct's slight drop in OCR (-1.75) may reflect \textbf{feedback overload}: overly granular corrections (e.g., ``Recount awning-tricycles after redefining `people''') could confuse smaller models with limited reasoning depth, leading to overcorrection.
\end{itemize}

\paragraph{Model Capacity Dictates Feedback Utilization}
The contrast between Pixtral-12B and Qwen2.5-VL-3B-Instruct underscores \textbf{scaling laws for feedback internalization}:
\begin{itemize}
    \item \textbf{Pixtral-12B} leverages structured feedback holistically, excelling in tasks requiring multi-step synthesis (e.g., OCR + counting). Its 58.25 score in OCR reflects an ability to chain corrections: first redefining terms (Feed-Target), then revising counts (Feed-Path).
    \item \textbf{Qwen2.5-VL-3B-Instruct} benefits most in \textbf{procedural tasks} (e.g., Instance Reasoning: +5.6) where feedback directly maps to executable steps. However, its performance plateaus in open-ended tasks (Logical Reasoning: +0.0), indicating limited capacity to generalize feedback beyond explicit instructions.
\end{itemize}

\paragraph{Feedback Type Impact}
Ablating feedback components would likely reveal hierarchical importance.
Notably, Direct Feedback's inconsistent results---sometimes trailing Standard Prompting (e.g., MMStar Logical Reasoning 42.4 vs. 48.8) suggest unstructured feedback introduces noise, confusing the student model with irrelevant or contradictory advice.

\paragraph{Error-Type Correctability}
Our method excels in correcting \textbf{systematic procedural errors} (e.g., misapplying definitions) but is less effective for \textbf{knowledge gaps}. For example, in Diagram tasks, feedback like ``Prioritize axis labels before interpreting trends'' directly resolves a common student error, whereas Logical Reasoning errors (e.g., flawed causality chains) require external knowledge beyond feedback's scope.

%
%
%
%

\begin{figure}
    \centering    \includegraphics[width=1.0\linewidth]{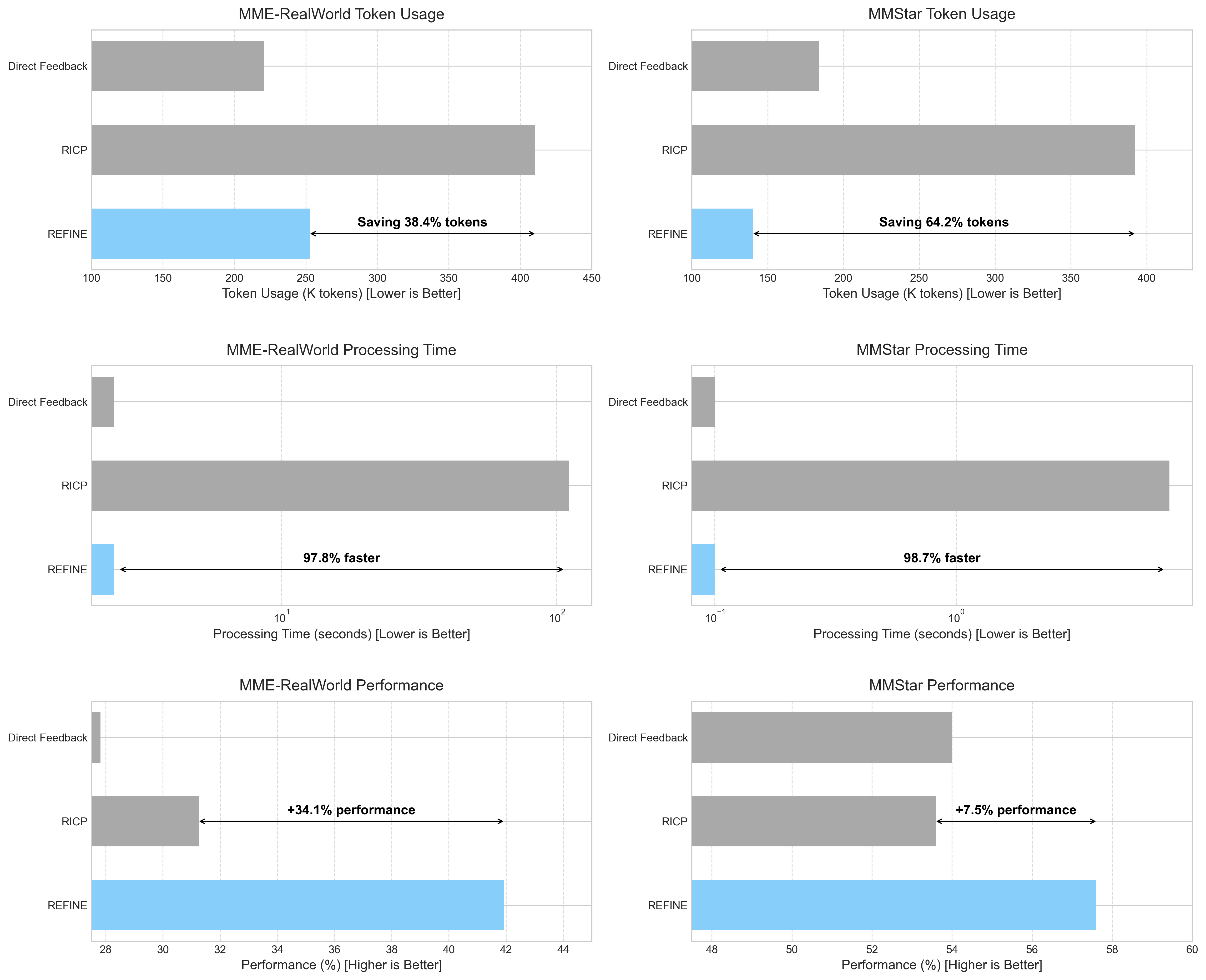}
    \caption{Performance (\%) of REFINE and baseline methods versus Token Usage and Processing Time.}
    \label{fig:comparison_results}
\end{figure}

\paragraph{Efficiency and Scalability of REFINE}
Besides accuracy, our method significantly outperformed baselines in terms of inference efficiency. Figure \ref{fig:comparison_results} shows that our structured feedback retrieval is substantially faster ($44.7-76.4\times$ speedup compared to RICP) and more token-efficient (approximately 64.2\% fewer tokens than RICP on MMStar). Precomputed embeddings and the removal of clustering significantly reduce computational costs, demonstrating scalability and feasibility in real-time settings. Furthermore, successful generalization from smaller subsets (MME-RealWorld-Lite) to larger-scale tests (MME-RealWorld) clearly illustrates the practical scalability of our approach.

\subsection{Ablation Study: Analyzing Feedback Component Contributions}
\begin{table}[ht]
\centering
\resizebox{0.5\textwidth}{!}{
\begin{tabular}{@{}lcc@{}}
\toprule
\textbf{Configuration} & \textbf{Overall} & \textbf{$\Delta$ from} \\
& \textbf{Score} & \textbf{Baseline} \\
\midrule
\textbf{Task/Process (Baseline)} & 41.92 & --- \\
+ Self-Reg                 & 32.50 & $-9.42$ ($-22.5\%$) \\
+ Cluster-Level            & 32.41 & $-9.51$ ($-22.7\%$) \\
+ CoT                      & 32.17 & $-9.75$ ($-23.3\%$) \\
+ Self-Reg + Cluster-Level & 32.07 & $-9.85$ ($-23.5\%$) \\
\bottomrule
\end{tabular}
}
\caption{Ablation study of REFINE's feedback components on MME-RealWorld (Overall Accuracy \%, Pixtral-12B). Shows performance impact when adding components to REFINE's core Task/Process feedback.}
\label{tab:components_analysis_results}
\end{table}

\paragraph{Feedback Components}
We conducted an ablation study by adding Cluster-level feedback and CoT.
\begin{itemize}
    \item \textbf{Cluster-level feedback}: To assess generalized feedback via clustering (cf. RICP~\citep{sun-etal-2024-retrieved}), we applied K-means (k=5) to the multimodal embeddings of training instances. For each data cluster, our teacher model generated a single generalized feedback from 20 samples of task/process feedback within that cluster. At inference, a query was assigned to its nearest cluster (by embedding similarity), combining this generalized feedback with REFINE's Task/Process feedback.
    \item \textbf{CoT}~\citep{wei2022chain}: Standard Chain-of-Thought prompting (`Let’s think step by step.') was appended after REFINE's Task/Process feedback to elicit explicit reasoning from the student model, rather than altering the feedback content itself.
    \item \textbf{Self-Reg}~\citep{hattie.timperley_2007}: Self-regulatory feedback, normally filtered out (Section~\ref{sec:feedback_filtering}), was re-introduced alongside Task/Process feedback.
\end{itemize}

To assess the impact of individual feedback types, we performed an ablation study (Table~\ref{tab:components_analysis_results}). The best overall performance (41.92) was achieved using Task/Process-level feedback exclusively, whereas the incorporation of additional feedback types consistently reduced model accuracy.

\paragraph{Specificity vs. Generalization Trade-off}
\textbf{Task/Process feedback} excels due to its direct alignment with the error context. For instance, advising the model to ``recount standing/walking poses'' directly addresses the miscounting error. But \textbf{Cluster-level feedback}, while intended to generalize insights, likely introduces noise. For example, feedback like ``check definitions in counting tasks'' may lack the precision needed for a \textbf{specific} question about ``people vs pedestrians,'' leading to ambiguous guidance.

The 22.7\% drop with Cluster-level feedback suggests that broad advice cannot substitute context-specific corrections.

\paragraph{Cognitive Overload in Multimodal Reasoning}
Adding \textbf{Self-Regulatory Feedback} (e.g., ``Reflect on past errors'') forces the model to split attention between executing the task and metacognitive monitoring, a challenge for vision-language models unoptimized for dual-task learning.

\textbf{CoT} exacerbates this by introducing open-ended reasoning steps (e.g., ``Think about object definitions'') that conflict with the structured corrective feedback. The 23.3\% drop with CoT highlights incompatibility between exploratory and directive instructions.

\paragraph{Compounding Noise in Combined Feedback}
The worst performance (-23.5\%) occurs when combining \textbf{Self-Regulatory + Cluster-level Feedback}. This suggests regulatory interference: the model receives vague Self-Regulatory cues \emph{and} overgeneralized cluster advice, diluting the actionable signal. For example, a prompt mixing ``Review past errors'' (Self-Regulatory) and ``Adjust counting strategies'' (Cluster) provides no concrete steps to correct a specific miscount.

\paragraph{Embedding-Clustering Mismatch}
The reliance on embeddings for clustering raises questions: if the embedding space fails to capture fine-grained task nuances (e.g., subtle differences in ``people'' definitions), clusters may group dissimilar errors, leading to irrelevant feedback retrieval.

\paragraph{Why Task/Process Feedback Works}
The success of Task/Process feedback aligns with principles of \emph{instructional alignment}:
\begin{itemize}
    \item \textbf{Goal-Oriented}: Directly answers ``What is needed to correct \emph{this} error?''
    \item \textbf{Procedural Clarity}: Provides stepwise actions (e.g., ``Reread the question, then recount'').
    \item \textbf{Minimal Abstraction}: Avoids meta-commentary, reducing cognitive load.
\end{itemize}
This approach mirrors effective human tutoring, where immediate, task-focused corrections yield better learning outcomes than abstract or generalized advice, a finding now validated for multimodal AI systems.
%
%
%

\section{Conclusion}
\label{sec:conclusion}

We introduced REFINE, a teacher-student framework that systematically structures errors to deliver targeted feedback for multimodal reasoning. REFINE employs three structured queries—Feed-Target, Feed-Check, and Feed-Path—to prioritize visual information, diagnose failures, and guide corrective actions. By optimizing structured feedback retrieval, unlike methods reliant on redundant retrievals, REFINE significantly improves inference speed, token efficiency, and scalability. Empirical results confirm substantial performance gains and robust generalization, highlighting REFINE's effectiveness in advancing multimodal reasoning.
\section*{Limitations}
The framework's effectiveness relies on the quality of teacher-generated feedback and the diversity of errors used in creating the Neural Error-book. Future research should explore its application to errors needing complex knowledge synthesis, its generalization to entirely new task domains, and further tailoring of feedback for varied model capacities.
\section*{Ethics Statement}
This research adhered to the ACL Ethics Policy, utilizing only publicly available datasets and large language models for evaluation. The work's goal is to investigate methods for enhancing reasoning, and no negative ethical outcomes are anticipated.

\bibliography{anthology_0,custom}

\clearpage 
\onecolumn
\appendix
\section{Dataset Statistics}
\begin{table}[h]
\begin{tabular}{lcc}
\toprule
\textbf{Benchmark} & \textbf{Error-book Size} & \textbf{Test Size} \\
\midrule
MME-RealWorld (Reasoning) & 534 & 2,092 \\
MMStar (Reasoning) & 123 & 250 \\
SEED-Bench-2-Plus & 563 & 1,141 \\
\bottomrule
\end{tabular}
\centering
\caption{Dataset split statistics for Error-book construction and testing.}
\label{tab:dataset_splits}
\end{table}
The Neural Error-book for each benchmark was constructed using only the questions the student model answered incorrectly during the training phase.

For the MME-RealWorld benchmark, the Error-book was created based on errors from the MME-RealWorld-Lite subset. 

For MMStar and SEED-Bench-2-Plus, the datasets were split into equal halves to create distinct train and test sets. The Error-book for these benchmarks was then built using the incorrectly answered questions from their respective 50\% train sets.
\section{REFINE Algorithm Detail}
\label{sec:appendix_algorithm}
\begin{algorithm}[ht]
\caption{REFINE}

\label{alg:code_refine}
\begin{algorithmic}[1]
\Require Image-question pairs $D$, Teacher model $T$, Student model $S$
\Ensure Enhanced predictions\\
\textbf{Note:} $T$ is a Teacher model generating three-stage structured feedback (Feed-Target/Check/Path).\\
\textbf{Note:} $\text{FeedbackFilter}$ filters to retain only actionable task/process-level feedbacks.

\State \textbf{Stage 1: Error Recognition and Insight Generation}
\State $D_{neg} \gets \emptyset$
\For{each $(q, img, ans_{gt}, ans_{pred}) \in D$ where $ans_{pred} \neq ans_{gt}$}
    \State $F_{target} \gets T(q, img, ans_{pred}, \text{"Define learning goals"})$ 
    \State $F_{check} \gets T(q, img, ans_{pred}, \text{"Check current progress"})$ 
    \State $F_{path} \gets T(q, img, ans_{pred}, \text{"Plan next steps"})$ 
    \State $F_{task/process} \gets \text{FeedbackFilter}(F_{target}, F_{check}, F_{path})$ 
    \State $D_{neg} \gets D_{neg} \cup \{(q, img, F_{task/proces})\}$
\EndFor

\State \textbf{Stage 2: Feedback Construction}
\For{each $(q, img, F_{task/process}) \in D_{neg}$}
    \State $e_{q,img} \gets \text{Embedding}(q, img)$
    \State $R \gets R \cup \{(e_{q,img}, F_{task/process})\}$
\EndFor

\State \textbf{Stage 3: Inference}
\For{each test pair $(q_{test}, img_{test})$}
    \State $e_{test} \gets \text{Embedding}(q_{test}, img_{test})$
    \State $F_{similar} \gets \text{NearestNeighbor}(e_{test}, R)$
    \State $ans \gets S(q_{test} + F_{similar}, img_{test})$ 
\EndFor
\end{algorithmic}
\end{algorithm}

\section{Prompt for REFINE}

\par\noindent\vspace{0.5em}

\begin{promptbox}{Prompt For Feed-Target}
Evaluate the student's response to a given question by reviewing their answer and providing feedback to reinforce the learning objective.
\\\\
Consider the scope and importance of the concept or skill that the student should grasp. Emphasize the relevance of this understanding in the broader context of the subject matter.
\\\\
\# Steps
\\\\
1. Review the original question and both the correct and incorrect answers provided by the student.

2. Identify the core learning objective or skill that the question aims to teach.

3. Clarify why this concept or skill is significant in the curriculum.

4. Relate the concept to the broader subject to illustrate its importance.
\\\\
\# Output Format
\\\\
Respond in a single sentence that concisely reiterates the primary learning goal, emphasizes its significance, and indicates what successful mastery looks like.
\\\\
\# Notes
\\\\\
- Ensure the feedback is encouraging and constructive, aiding student comprehension and motivation.

- Tailor the response to align with educational standards and learning objectives specific to the curriculum.
\\\\
IMPORTANT: DO NOT directly mention the correct answer or the corresponding options.
\\\\
\texttt{Question: \{A specific question from the dataset\}} \\
\texttt{Correct Answer: \{The ground-truth answer for the question\}} \\
\texttt{Incorrect student's Answer: \{The model's incorrect answer\}}
\end{promptbox}

\vspace{2em} 

\begin{promptbox}{Prompt For Feed-Check}
Review the student's solution or explanation and evaluate the correctness of their approach by identifying correct elements, areas showing partial understanding, and any misunderstandings or errors.
\\\\
Focus on pinpointing both accurate and inaccurate components, clearly articulating any misunderstandings such as conceptual errors, calculation slips, or instruction misinterpretations. Use precise language to highlight the analysis, such as noting specific steps where mistakes occur.
\\\\
\# Output Format
\\\\
Provide your evaluation in a single, clear, and specific sentence that encompasses both strengths and misconceptions in the student's response.
\\\\
\# Notes
\\\\
- Ensure to maintain the single-sentence structure for clarity and brevity.\\
- Clearly separate each element of the evaluation (correct parts, partial understanding, errors) within the sentence.\\ 
- Emphasize spotting both strengths and areas in need of improvement.
\\\\
IMPORTANT: DO NOT directly mention the correct answer or the corresponding options.
\\\\
\texttt{Question: \{A specific question from the dataset\}} \\
\texttt{Correct Answer: \{The ground-truth answer for the question\}} \\
\texttt{Incorrect student's Answer: \{The model's incorrect answer\}}
\end{promptbox}

\vspace{2em}

\begin{promptbox}{Prompt For Feed-Path}
Review the question, the correct answer, and the incorrect answer provided by the model, then respond with actionable strategies to guide the student in correcting their misunderstanding and improving their understanding.
\\\\
Offer both short-term and long-term steps, focusing on concrete, constructive actions that can help the student progress.
\\\\
- Encourage actionable, bite-sized advice.\\
- Suggest multiple methods for approaching the subject.\\
- Maintain a supportive tone, emphasizing that mistakes are a learning opportunity.
\\\\
\# Output Format
\\\\
Provide your response in a single sentence, integrating specific actions the student can take to address their misunderstanding, and strategies to build on their correct reasoning and fix their errors.
\\\\
IMPORTANT: DO NOT directly mention the correct answer or the corresponding options.
\\\\
\texttt{Question: \{A specific question from the dataset\}} \\
\texttt{Correct Answer: \{The ground-truth answer for the question\}} \\
\texttt{Incorrect student's Answer: \{The model's incorrect answer\}}
\end{promptbox}


\section{Detailed Example of REFINE}

\begin{tcolorbox}[
    enhanced,
    breakable,
    colback=gray!10,
    colframe=gray!75,
    fonttitle=\large\bfseries,
    coltitle=black,
    title={\textbf{REFINE in Action} \quad \faImage},
    boxrule=1pt,
    arc=3mm,
    left=6mm, right=6mm, top=4mm, bottom=6mm
    ]

    \includegraphics[width=0.9\linewidth]{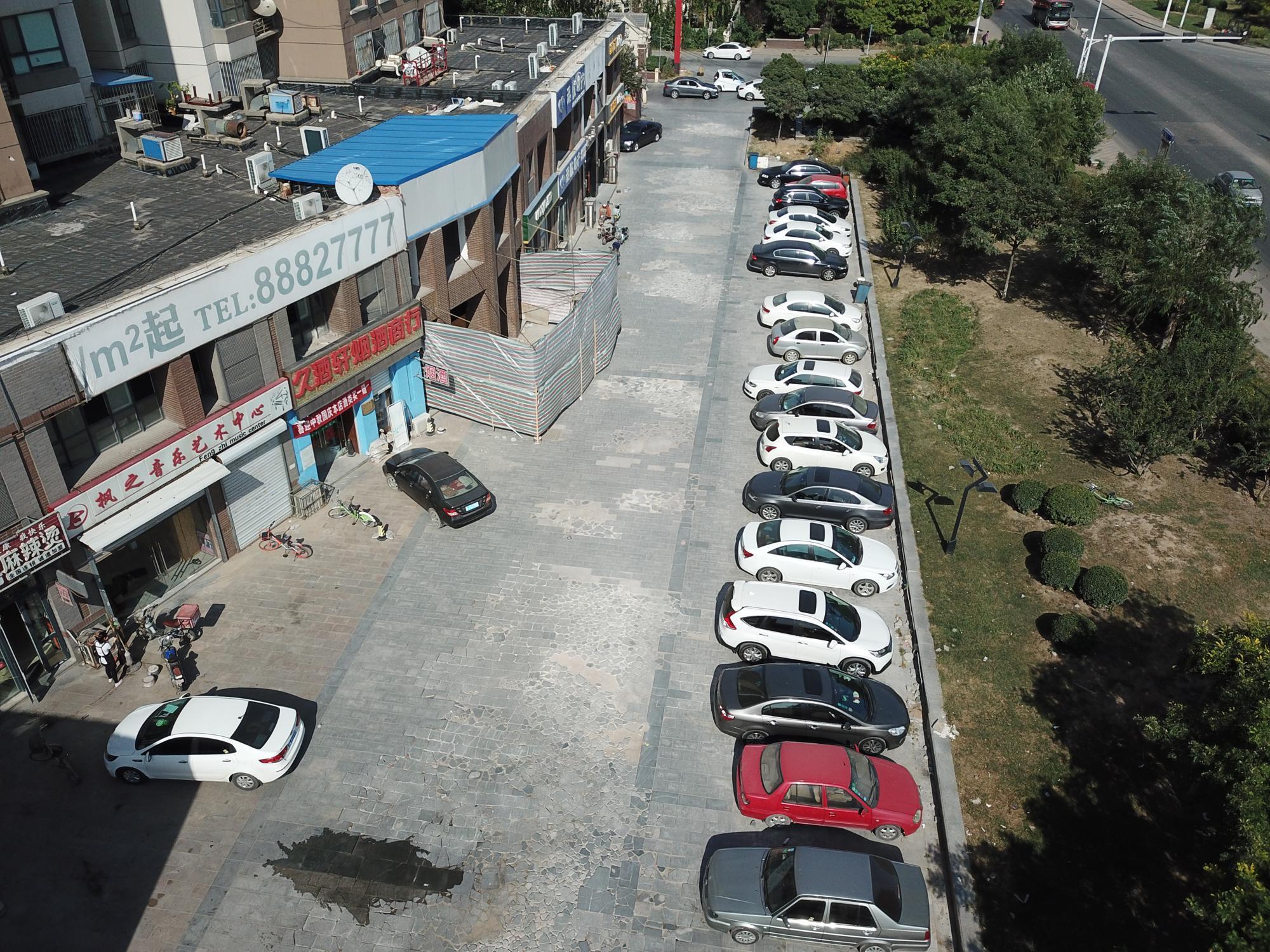}
    \label{fig:refine_example}
    \vspace{4mm}
    
    \boxheading{Question}
    
    What is the total number of people and awning-tricycles in the image? (If a human maintains standing pose or walking, please classify it as pedestrian, otherwise, it is classified as a people.)
    
The choices are listed below:

(A) 15

(B) 4

(C) 0

(D) 3

(E) The image does not feature the objects

    \boxheading{Model Response}
    
    The image shows a street scene with parked cars, buildings, and a few pedestrians. There are no awning-tricycles visible in the image. To determine the total number of people and awning-tricycles, we need to count the number of people in the image. Upon examining the image, \textbf{we can see a few pedestrians walking on the sidewalk. Specifically, there are three people visible} in the image. Since there are no awning-tricycles in the image, the total number of people and awning-tricycles is simply the number of people, which is 3.
    \par\noindent\textbf{Final Answer: D} 
    \textcolor{boxWrongRed}{\textbf{[WRONG]}}
    
    \tcblower 

    \boxheading{Retrieved Task/Process-level Feedback (from a similar past error)}

    \begin{tcolorbox}[
        enhanced,
        colback=feedbackYellow,
        colframe=orange!80!black,
        boxrule=0.5pt, arc=1mm,
        left=3mm, right=3mm, top=3mm, bottom=3mm
        ]
        During solving the solution, refer to the following information: \\
        The student correctly identifies two awning-tricycles, but misinterprets the definition of "people," counting 10 individuals when only two meet the specified criteria of not being in a standing or walking pose, leading to an incorrect total of 12 instead of 2. \textbf{Reread the question carefully, noting the specific definition of "people" provided, then review the image focusing on individuals in standing or walking poses} and awning-tricycles, and finally recount these specific objects to arrive at the correct total.
    \end{tcolorbox}

    \boxheading{REFINE Response}
    
    The image shows a street scene with parked cars, buildings, and a few people. There are no awning-tricycles visible in the image.
    \par Let's analyze the image step-by-step:
    \begin{enumerate}[leftmargin=*, itemsep=1pt, topsep=2pt]
        \item Identify the objects in the image: The image shows a street with parked cars, a few people, and buildings. There are no awning-tricycles visible.
        \item Count the number of people: There are a few people visible in the image, \textbf{but we need to count only those who are not in a standing or walking pose.}
        \item Count the number of awning-tricycles: There are no awning-tricycles visible in the image.
    \end{enumerate}
    Based on the analysis, the total number of people and awning-tricycles in the image is 0.
    \par\noindent\textbf{Final Answer: C} \textcolor{boxCorrectGreen}{\textbf{[CORRECT]}}

\end{tcolorbox}

\end{document}